\documentclass[final,nonatbib]{article}
\usepackage{neurips_2021}
\usepackage[utf8]{inputenc} 
\usepackage[T1]{fontenc}    
\usepackage{booktabs}       
\usepackage{amsfonts}
\usepackage{nicefrac}
\usepackage{microtype}      
\usepackage{xcolor,bm,tikz,amssymb,amsthm,pgfplots,fourier,subfigure,algorithm,algorithmic,amsmath,blkarray,todonotes,xparse,soul,comment,url,hyperref}
 
\usetikzlibrary{arrows,automata,circuits}
\usetikzlibrary{arrows,plotmarks,decorations.markings,trees,shapes}
\tikzstyle{nodo}=[ellipse,draw=black!100,fill=black!0,line width=.7pt,minimum width=1.0cm,minimum height=0.8cm,
text width=1cm,text centered]
\tikzstyle{nodo2}=[ellipse,draw=black!100,fill=black!10,line width=.7pt,minimum width=1.0cm,minimum height=0.8cm,
text width=1cm,text centered]
\tikzstyle{nodo3}=[ellipse,draw=black!100,fill=black!30,line width=.7pt,minimum width=1.0cm,minimum height=0.8cm,
text width=1cm,text centered]
\tikzstyle{Qnodo}=[ellipse,draw=black!100,fill=black!10,line width=.7pt,minimum width=1.2cm,minimum height=.7cm]
\tikzstyle{arco}=[draw=black!80,line width=.7pt, postaction={decorate}, decoration={markings,mark=at position 1.0 with {\arrow[ draw=black!80,line width=.7pt]{>}}}]
\tikzstyle{decision} = [rectangle, draw, fill=black!100,text=white, text width=4.5em, text badly centered, node distance=3cm, minimum height=3em]
\tikzstyle{block} = [rectangle, draw, fill=blue!20, text width=5em, text centered, rounded corners, minimum height=3em]
\tikzstyle{line} = [draw, -latex']
\tikzstyle{cloud} = [draw, ellipse,fill=red!20, node distance=3cm, minimum height=2em]

\pgfplotsset{legend image with text/.style={
legend image code/.code={%
\node[anchor=center] at (0.3cm,0cm) {#1};}},}
\pgfplotsset{compat=1.17}
\usetikzlibrary{arrows,plotmarks,decorations.markings,trees,shapes,pgfplots.groupplots}
\tikzset{dot/.style = {circle, fill, minimum size=#1,inner sep=0pt, outer sep=0pt, fill, circle},dot/.default = 6pt}
\tikzset{dot2/.style = {circle, fill, color=black!40,minimum size=6pt,inner sep=0pt, outer sep=0pt, fill, circle}}
\tikzstyle{a}=[->,>=stealth,dashed]
\tikzstyle{a2}=[->,>=stealth]
\tikzstyle{a3}=[<->,>=stealth]

\newtheorem{thm}{Theorem}
\newtheorem{cor}{Corollary}

\newtheorem{exe}{Example}
\newtheorem{prf}{Proof of Theorem}

\title{Causal Expectation-Maximisation}
\author{%
Marco Zaffalon\\
IDSIA (Switzerland)\\
\texttt{zaffalon@idsia.ch}\\
\And
Alessandro Antonucci\\
IDSIA (Switzerland)\\
\texttt{alessandro@idsia.ch}\\
\And Rafael Caba\~nas\\
IDSIA (Switzerland)\\
\texttt{rcabanas@idsia.ch}}
\begin{document}
\maketitle
\begin{abstract}
Structural causal models are the basic modelling unit in Pearl's causal theory; in principle they allow us to solve counterfactuals, which are at the top rung of the ladder of causation. But they often contain latent variables that limit their application to special settings. This appears to be a consequence of the fact, proven in this paper, that causal inference is NP-hard even in models characterised by polytree-shaped graphs. To deal with such a hardness, we introduce the causal EM algorithm. Its primary aim is to reconstruct the uncertainty about the latent variables from data about categorical manifest variables. Counterfactual inference is then addressed via standard algorithms for Bayesian networks. The result is a general method to approximately compute counterfactuals, be they identifiable or not (in which case we deliver bounds). We show empirically, as well as by deriving credible intervals, that the approximation we provide becomes accurate in a fair number of EM runs. These results lead us finally to argue that there appears to be an unnoticed limitation to the trending idea that counterfactual bounds can often be computed without knowledge of the structural equations.
\end{abstract}
\maketitle

\section{Introduction}\label{sec:intro}
\begin{figure}[htp!]
\centering
\begin{subfigure}{(a)}
\begin{tikzpicture}[scale=0.65]
\node[dot,label=left:{$X$}] (x)  at (0,0) {};
\node[dot,label=right:{$Y$}] (y)  at (1,0) {};
\draw[a2] (x) -- (y);
\end{tikzpicture}
\end{subfigure}
\begin{subfigure}{(b)}
\begin{tikzpicture}[scale=0.65]
\node[dot,label=left:{$X$}] (x)  at (0,0) {};
\node[dot,label=right:{$Y$}] (y)  at (1,0) {};
\node[dot2,label=left:{$U$}] (u)  at (0,1) {};
\node[dot2,label=right:{$V$}] (v)  at (1,1) {};
\draw[a] (u) -- (x);
\draw[a] (v) -- (y);
\draw[a2] (x) -- (y);
\end{tikzpicture}
\end{subfigure}
\begin{subfigure}{(c)}
\begin{tikzpicture}[scale=0.65]
\node[dot,label=left:{$X$}] (x)  at (0,0) {};
\node[dot,label=right:{$Y$}] (y)  at (1,0) {};
\node[dot,label=left:{$X'$}] (x1)  at (0,2) {};
\node[dot,label=right:{$Y'$}] (y1)  at (1,2) {};
\node[dot2,label=left:{$U$}] (u)  at (0,1) {};
\node[dot2,label=right:{$V$}] (v)  at (1,1) {};
\draw[a] (u) -- (x);
\draw[a] (v) -- (y);
\draw[a] (v) -- (y1);
\draw[a] (u) -- (x1);
\draw[a2] (x) -- (y);
\draw[a2] (x1) -- (y1);
\end{tikzpicture}
\end{subfigure}
\begin{subfigure}{(d)}
\begin{tikzpicture}[scale=0.65]
\node[dot,label=left:{$X$}] (x)  at (0,0) {};
\node[dot,label=right:{$Y$}] (y)  at (1,0) {};
\node[dot,label=left:{$X'$}] (x1)  at (0,2) {};
\node[dot,label=right:{$Y'$}] (y1)  at (1,2) {};
\node[dot2,label=left:{$U$}] (u)  at (0,1) {};
\node[dot2,label=right:{$V$}] (v)  at (1,1) {};
\draw[a] (u) -- (x);
\draw[a] (v) -- (y);
\draw[a] (v) -- (y1);
\draw[a2] (x) -- (y);
\draw[a2] (x1) -- (y1);
\end{tikzpicture}
\end{subfigure}\\
\begin{subfigure}{(e)}
\begin{tikzpicture}[scale=0.8]
\node[dot,label=left:{$X$}] (x)  at (0,0) {};
\node[dot,label=above right:{$Y$}] (y)  at (2,0) {};
\node[dot,label=above right:{$Z$}] (z)  at (1,1) {};
\draw[a2] (x) -- (y);
\draw[a2] (z) -- (x);
\draw[a2] (z) -- (y);
\end{tikzpicture}
\end{subfigure}
\begin{subfigure}{(f)}
\begin{tikzpicture}[scale=0.65]
\node[dot,label=left:{$X$}] (x)  at (0,0) {};
\node[dot,label=above right:{$Y$}] (y)  at (2,0) {};
\node[dot,label=above right:{$Z$}] (z)  at (1,1) {};
\node[dot2,label=left:{$U$}] (u)  at (0,1) {};
\node[dot2,label=left:{$W$}] (w)  at (1,2) {};
\node[dot2,label=right:{$V$}] (v)  at (2,1) {};
\draw[a] (u) -- (x);
\draw[a] (w) -- (z);
\draw[a] (v) -- (y);
\draw[a2] (x) -- (y);
\draw[a2] (z) -- (x);
\draw[a2] (z) -- (y);
\end{tikzpicture}
\end{subfigure}
\begin{subfigure}{(g)}
\begin{tikzpicture}[scale=0.65]
\node[dot,label=left:{$X$}] (x)  at (0,0) {};
\node[dot,label=above right:{$Y$}] (y)  at (2,0) {};
\node[dot,label=above right:{$Z$}] (z)  at (1,1) {};
\node[dot,label=below right:{$Z'$}] (z1)  at (1,3) {};
\node[dot,label=below right:{$Y'$}] (y1)  at (2,4) {};
\node[dot,label=left:{$X'$}] (x1)  at (0,4) {};
\node[dot2,label=left:{$U$}] (u)  at (0,2) {};
\node[dot2,label=left:{$W$}] (w)  at (1,2) {};
\node[dot2,label=left:{$V$}] (v)  at (2,2) {};
\draw[a] (u) -- (x);
\draw[a] (w) -- (z);
\draw[a] (v) -- (y);
\draw[a] (u) -- (x1);
\draw[a] (w) -- (z1);
\draw[a] (v) -- (y1);
\draw[a2] (x) -- (y);
\draw[a2] (z) -- (x);
\draw[a2] (z) -- (y);
\draw[a2] (x1) -- (y1);
\draw[a2] (z1) -- (x1);
\draw[a2] (z1) -- (y1);
\end{tikzpicture}
\end{subfigure}
\begin{subfigure}{(h)}
\begin{tikzpicture}[scale=0.65]
\node[dot,label=left:{$X$}] (x)  at (0,0) {};
\node[dot,label=above right:{$Y$}] (y)  at (2,0) {};
\node[dot,label=above right:{$Z$}] (z)  at (1,1) {};
\node[dot,label=below right:{$Z'$}] (z1)  at (1,3) {};
\node[dot,label=below right:{$Y'$}] (y1)  at (2,4) {};
\node[dot,label=left:{$X'$}] (x1)  at (0,4) {};
\node[dot2,label=left:{$W$}] (w)  at (1,2) {};
\node[dot2,label=left:{$V$}] (v)  at (2,2) {};
\draw[a] (w) -- (z);
\draw[a] (v) -- (y);
\draw[a] (w) -- (z1);
\draw[a] (v) -- (y1);
\draw[a2] (x) -- (y);
\draw[a2] (z) -- (x);
\draw[a2] (z) -- (y);
\draw[a2] (x1) -- (y1);
\draw[a2] (z1) -- (x1);
\draw[a2] (z1) -- (y1);
\end{tikzpicture}
\end{subfigure}
\caption{A causal diagram over two variables (a), a corresponding (Markovian) SCM over those two (endogenous) variables (b), its counterfactual graph (c), the result of a surgery to compute PN (d), and another causal diagram over three variables (e), the corresponding (still Markovian) SCM (f), its counterfactual graph (g), and the result of a surgery to compute PNS (h).
\label{fig:scmtoy}}
\end{figure}
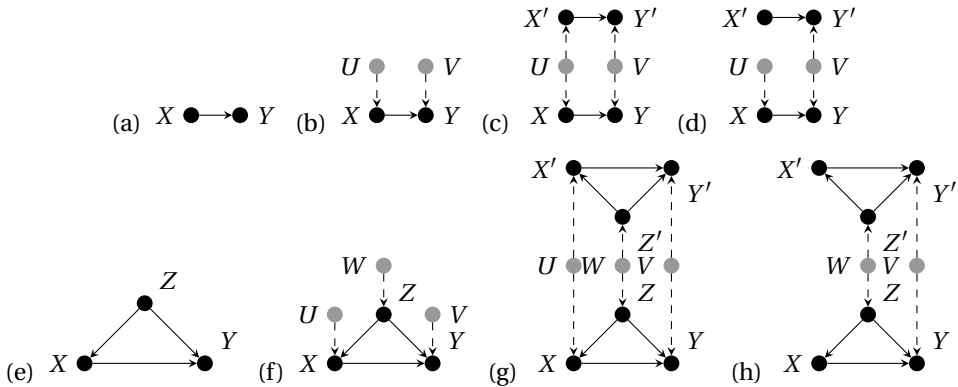

Structural causal models (SCMs) are central to Pearl's theory of causal inference. An SCM can be displayed graphically as in Fig.~\ref{fig:scmtoy}b, where $U$ and $V$ represent \emph{exogenous} variables. By definition, they have an associated distribution; internal, \emph{endogenous}, nodes are instead deterministic functions of their parents. These define so-called \emph{structural equations} (SE), which assign values to endogenous nodes. Exogenous variables are often latent in SCMs, while we usually have data about the endogenous ones. In this case we talk of a \emph{partially specified} SCM (PSMC); if we have data also for the exogenous variables, then we say that the SCM is \emph{fully specified} (FSCM).

FSCMs permit doing counterfactual inference, which answers hypothetical questions like, what if I had done something differently in the past? This type of inference is strictly more general than computing interventions (what if I do this?), which is strictly more general than probabilistic reasoning (what if I see this?).\footnote{Except for sets of measure zero, see \cite{bareinboim2020pearl}.} Computing probabilities of counterfactuals is therefore of utmost importance in causal inference, and one can achieve it by an FSCM.

PSCMs allow in principle to compute those probabilities too; however, most often the result is non-identifiable, which means that those probabilities can only be known to belong to an interval. Having interval-valued outcomes is often useful, in spite of their inherent imprecision. But only few methods are available to compute non-identifiable quantities, and most of them are ad hoc. Three approaches stand out as more systematic: the first, from 2020, converts the problem into one of inference in `credal networks' and uses algorithms for these to yield exact counterfactual intervals \cite{zaffalon2020}. The method is efficient on Markovian SCMs and can solve a class of non-Markovian ones too, albeit with increasing complexity. The second, from 2021, converts the problem into one of polynomial optimisation and, given its complexity, eventually approximates its solution via Monte Carlo sampling \cite{zhang2021}. However it does not present an extensive empirical evaluation, so that it is unclear how accurate vs. time consuming is the approximation. The third, again from 2021, is similar to the previous one \cite{duarte2021}. The difference is that the polynomial optimisation is used to yield exact solutions in some case. Approximations are provided for the general case; extensive empirical evaluations are missing.

In this paper we present a general method to compute approximate intervals of counterfactual probabilities. We do this after observing that the computation of counterfactual probabilities is NP-hard even in polytree topologies of PSCM; remember that in FSCMs the computation is polynomial. This shows that PSCMs are particularly complex models to handle.

The solution method we propose can be understood as a specialisation of the EM algorithm to causal problems. We show that in the case of PSCMs, the EM always converges to a global optimum: that is, a distribution for the exogenous variables that generates, through the given SCM, the available empirical probabilities; we call this a \emph{compatible} distribution. Multiple EM runs will amount to sampling the set of compatible distributions, or, stated differently, to sampling multiple FSCMs compatible with the given PSCM. Taking the extrema of the counterfactual probabilities that those FSCMs yield, we obtain an interval that is included in the original PSCM's actual interval.

Our experiments show that 20 EM runs are enough to obtain a good approximation to the actual interval. We double-check this result by analytically developing credible bounds: these indeed show that the approximate interval is close to the actual one with 95\% credibility just after 20 runs.

In the special, and especially important, case where the outcome is identifiable, 6 EM runs are enough to be that confident that it is indeed identifiable (9 runs for 99\% credibility). Remember that Pearl's celebrated do-calculus permits, in principle, computing the probability of the effect of interventions, when the former is identifiable. In such a case, our EM-based algorithm can be regarded as a very viable numerical extension of do-calculus to counterfactuals---and without the need to handle the algebraic manipulations of do-calculus, since it yields solutions automatically.

Finally, we discuss an issue that does not seem to have been noticed so far: that counterfactual bounds are unwarranted if obtained in absence of a pre-defined PSCM of the problem. Stated differently, in general one cannot give bounds that hold irrespectively of the actual underlying PSCM; structural equations can hardly be neglected.

Outline of the paper: Sect.~\ref{sec:background} introduces SCMs. Sect.~\ref{sec:lik} characterises the set of compatible distributions and defines SCM-compatibility. Sect.~\ref{sec:em} proves that computation in PSCMs is NP-hard, presents the causal EM and proves its optimality. Sect.~\ref{sec:incomp} discusses the problems we meet when we neglect structural equations. Sect.~\ref{sec:exp} presents the credible intervals and the experiments. Sect.~\ref{sec:conc} concludes the paper. Proofs are gathered in Appendix~\ref{sec:app}.

\section{Structural Causal Models}\label{sec:background}
We focus on models with categorical variables.\footnote{Exogenous variables can be assumed categorical w.l.o.g. if endogenous ones are (\cite[Prop.~1]{duarte2021}; \cite[Thm.~1]{zhang2021}).} Variable $V$ takes values from a finite set $\Omega_V$; $v$ denotes its generic value. $P(V)$ is a probability mass function (PMF) over $V$. Notation $P(V|V'):=\{P(V|v')\}_{v'\in\Omega_{V'}}$ is used for a conditional probability table, i.e., a collection of conditional PMFs over a variable indexed by the states of another one. We define an SCM $M$ by a directed graph $\mathcal{G}$ whose nodes are in a one-to-one correspondence with both its endogenous variables $\bm{X}:=(X_1,\ldots,X_n)$ and the exogenous ones $\bm{U}:=(U_1,\ldots,U_m)$. We focus on semi-Markovian models, i.e., $\mathcal{G}$ is assumed acyclic. Exogenous variables correspond to the root nodes of $\mathcal{G}$. A PMF $P(U)$ is specified for each $U\in\bm{U}$. A structural equation (SE) $f_X$ is instead provided for each $X \in \bm{X}$; it is a map $f_X:\Omega_{\mathrm{Pa}_X} \to \Omega_X$, where $\mathrm{Pa}_X$ are the parents (i.e., the immediate predecessors) of $X$ according to $\mathcal{G}$. If $X$ has a single exogenous parent with states indexing all the possible deterministic relations between the endogenous parents and $X$, we call the specification of $f_X$ \emph{conservative}. To have all the states of $X$ possibly realised, we only consider surjective SEs. We also require a joint surjectivity, meaning that any $\bm{x}\in\Omega_{\bm{X}}$ can be realised for at least a $\bm{u}\in\Omega_{\bm{U}}$. SCM $M$ induces the following joint PMF:
\begin{equation}\label{eq:joint}
P(\bm{x},\bm{u})=\prod_{X\in\bm{X}} P(x|\mathrm{pa}_X) \prod_{U\in\bm{U}} P(u)\,,
\end{equation}
for each $\bm{x}\in\Omega_{\bm{X}}$ and $\bm{u}\in\Omega_{\bm{U}}$, where the values of the conditional probability tables associated with each $X$ are degenerate, i.e.,
$P(x|\mathrm{pa}_X)=\delta_{x,f_X(\mathrm{pa}_X)}$. These are fully specified SCMs (i.e., FSCM). If the exogenous PMFs are not provided, we say instead that $M$ is partially specified (PSCM). In that case, information about $\bm{X}$ is assumed to be available, e.g., in a data set $\mathcal{D}$ of observations.

\section{Likelihood Characterisation}\label{sec:lik}
The factorisation properties of the joint PMF $P(\bm{X},\bm{U})$ of an SCM $M$, as shown in Eq.~\eqref{eq:joint}, are those of a Bayesian network (BN) defined over graph $\mathcal{G}$. A characterisation can be provided also for the marginal graph related to the joint $P(\bm{X})$ via the notion of c-component \cite{tian2002studies} (see also Ex.~\ref{ex:primo} below.) 

Given the graph $\mathcal{G}$ of $M$, its \emph{reduction} $\mathcal{G}'$ is obtained by removing from $\mathcal{G}$ any arc connecting pairs of endogenous variables (e.g., see Fig.~\ref{fig:cc}b). Let $\{\mathcal{G}_c\}_{c\in\mathcal{C}}$ denote the connected components of $\mathcal{G}'$. The \emph{c-components} of $M$ are the elements of the partition $\{\bm{X}^{(c)}\}_{c\in\mathcal{C}}$ of $\bm{X}$, where $\bm{X}^{(c)}$ denotes the endogenous nodes of $\mathcal{G}_c$ for each $c\in\mathcal{C}$. This definition is consistent with the one in \cite{tian2002studies}, where the more general case of possibly non-root exogenous nodes is considered. We similarly denote as $\{\bm{U}^{(c)}\}_{c\in\mathcal{C}}$ the partition of $\bm{U}$ induced by the connected components of $\mathcal{G}'$. For each $c\in\mathcal{C}$, we denote as $\bm{Y}^{(c)}$ the union of the endogenous parents of the nodes in $\bm{X}^{(c)}$ and $\bm{X}^{(c)}$ itself. Finally, for each $X\in \bm{X}^{(c)}$,  $\bm{Y}^{(c)}_X$ is obtained by removing from $\bm{Y}^{(c)}$ the nodes topologically following $X$ and $X$ itself. The c-components induce the following endogenous factorisation \cite[Chp.~4, Cor.~4]{tian2002studies}:
\begin{equation}\label{eq:empirical}
P(\bm{x})=\prod_{c \in \mathcal{C}} \prod_{X\in\bm{X}^{(c)}} P(x|\bm{y}^{(c)}_X)\,,
\end{equation}
for each $\bm{x}\in\Omega_{\bm{X}}$, with the values of $x$ and $\bm{y}^{(c)}_X$ consistent with $\bm{x}$. An example of a graph corresponding to such a factorisation is in Fig.~\ref{fig:cc}c. We can also write a factorisation with respect to the c-components in Eq.~\eqref{eq:joint}, that is associated with the original graph (e.g., see Fig.~\ref{fig:cc}a), as follows:
\begin{equation}\label{eq:joint2}
P(\bm{x},\bm{u}) =
\prod_{c\in\mathcal{C}}
\left[
\prod_{X\in\bm{X}^{(c)}} P(x|\mathrm{pa}_X)
\prod_{U\in\bm{U}^{(c)}} P(u)
\right]\,.
\end{equation}
\begin{exe}\label{ex:primo}
Fig.~\ref{fig:cc}a depicts an SCM with three c-components, say $\mathcal{C}=\{1,2,3\}$. As shown by Fig.~\ref{fig:cc}b: $\bm{X}^{(1)}=\{X_1\}$,
 $\bm{X}^{(2)}=\{X_2,X_4\}$, $\bm{X}^{(3)}=\{ X_3\}$, $\bm{U}^{(c)}=\{ U_c\}$ for each $c\in\mathcal{C}$, $\bm{Y}^{(1)}=\{X_1\}$, $\bm{Y}^{(1)}_{X_1}=\emptyset$,
 $\bm{Y}^{(2)}=\{X_1,X_2,X_3,X_4\}$,
 $\bm{Y}^{(2)}_{X_2}=\{X_1\}$,
 $\bm{Y}^{(2)}_{X_4}=\{X_1,X_2,X_3\}$, and
 $\bm{Y}^{(3)}=\{X_2,X_3\}$ and $\bm{Y}^{(3)}_{X_3}=\{X_2\}$.
 Eq.~\eqref{eq:empirical} gives therefore the endogenous factorisation induced by the graph in Fig.~\ref{fig:cc}c. 
\end{exe}
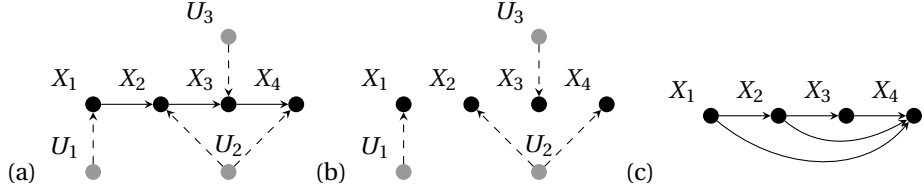
\begin{figure}[htp!]
\centering
\begin{subfigure}{(a)}
\begin{tikzpicture}[scale=0.9]
\node[dot,label=above left:{$X_1$}] (x1)  at (0,0) {};
\node[dot2,label=above left:{$U_1$}] (u1)  at (0,-1) {};
\node[dot2,label=above:{$U_2$}] (u2)  at (2,-1) {};
\node[dot2,label=above left:{$U_3$}] (u3)  at (2,1) {};
\node[dot,label=above left:{$X_2$}] (x2)  at (1,0) {};
\node[dot,label=above left:{$X_3$}] (x3)  at (2,0) {};
\node[dot,label=above left:{$X_4$}] (x4)  at (3,0) {};
\draw[a] (u1) -- (x1);
\draw[a] (u3) -- (x3);
\draw[a] (u2) -- (x2);
\draw[a] (u2) -- (x4);
\draw[a2] (x1) -- (x2);
\draw[a2] (x2) -- (x3);
\draw[a2] (x3) -- (x4);
\end{tikzpicture}
\end{subfigure}
\begin{subfigure}{(b)}
\begin{tikzpicture}[scale=0.9]
\node[dot,label=above left:{$X_1$}] (x1)  at (0,0) {};
\node[dot2,label=above left:{$U_1$}] (u1)  at (0,-1) {};
\node[dot2,label=above:{$U_2$}] (u2)  at (2,-1) {};
\node[dot2,label=above left:{$U_3$}] (u3)  at (2,1) {};
\node[dot,label=above left:{$X_2$}] (x2)  at (1,0) {};
\node[dot,label=above left:{$X_3$}] (x3)  at (2,0) {};
\node[dot,label=above left:{$X_4$}] (x4)  at (3,0) {};
\draw[a] (u1) -- (x1);
\draw[a] (u3) -- (x3);
\draw[a] (u2) -- (x2);
\draw[a] (u2) -- (x4);
\end{tikzpicture}
\end{subfigure}
\begin{subfigure}{(c)}
\begin{tikzpicture}[scale=0.9]
\node[dot,label=above left:{$X_1$}] (x1)  at (0,0) {};
\node[dot,label=above left:{$X_2$}] (x2)  at (1,0) {};
\node[dot,label=above left:{$X_3$}] (x3)  at (2,0) {};
\node[dot,label=above left:{$X_4$}] (x4)  at (3,0) {};
\draw[a2] (x1) -- (x2);
\draw[a2] (x2) -- (x3);
\draw[a2] (x3) -- (x4);
\draw[a2] (x1) [out=-40,in=-130] to (x4);
\draw[a2] (x2) [out=-40,in=-150] to (x4);
\end{tikzpicture}
\end{subfigure}
\caption{A SCM (a), its reduction (b), its endogenous factorisation (c).
\label{fig:cc}}
\end{figure}

Notice that Markovianity \cite{pearl2009causality} corresponds to condition
$|\bm{X}^{(c)}|=|\bm{U}^{(c)}|=1$ for each $c \in \mathcal{C}$. Having only $|\bm{U}^{(c)}|=1$ for each $c\in\mathcal{C}$ is called instead quasi-Markovianity \cite{zaffalon2020}.

Now consider an FSCM $M$, from which we sample a set $\mathcal{D}$ of i.i.d. endogenous observations. As a consequence of the factorisation in Eq.~\eqref{eq:empirical}, the log-likelihood of $\bm{x}$ given $\mathcal{D}$ is:
\begin{equation}\label{eq:likdec}
LL[\{P(x|\bm{y}_X^{(c)})\}]:= \sum_{\bm{x} \in\mathcal{D}} \sum_{c\in\mathcal{C}} \sum_{X\in\bm{X}^{(c)}} \log P(x|\bm{y}^{(c)}_X)
=
\sum_{c\in\mathcal{C}} \sum_{X\in\bm{X}^{(c)}} 
\sum_{x,\bm{y}_X^{(c)}}
n(x,\bm{y}^{(c)}_X) \cdot
\log P(x|\bm{y}^{(c)}_X) 
\,,
\end{equation}
where the values of $x$ and $\bm{y}_X^{(c)}$ are those consistent with $\bm{x}$, $n(\cdot)$ denote the frequencies in $\mathcal{D}$ of its argument, and indexes in the argument of LL are omitted for the sake of readability. In the rightmost-hand side of Eq.~\eqref{eq:likdec}, we see the decomposable structure of a multinomial likelihood. Such a concave function has no local maxima and a global maximum value given by 
\begin{equation}\label{eq:LL*}
LL^*:= \sum_{c\in\mathcal{C}} \sum_{X\in\bm{X}^{(c)}} 
\sum_{x,\bm{y}_X^{(c)}}
n(x,\bm{y}^{(c)}_X) \cdot
\log\hat{P}(x|\bm{y}^{(c)}_X) 
\,,
\end{equation}
which is achieved if and only if the probabilities attain the relative frequencies in $\mathcal{D}$,  i.e.,
\begin{equation}\label{eq:freqs}
\hat{P}(x|\bm{y}_X^{(c)}):=\frac{n(x,\bm{y}_X^{(c)})}{n(\bm{y}_X^{(c)})}\,,
\end{equation}
for each $x$ and $\bm{y}_X^{(c)}$, $c\in\mathcal{C}$, and $X\in\bm{X}^{(c)}$  \cite{koller2009}.

Note that in the limit of infinite data, the endogenous PMF we obtain from the frequencies coincides with the marginalisation of the exogenous variables from the joint PMF in Eq.~\eqref{eq:joint2}, i.e.,
\begin{equation}\label{eq:compat}
\prod_{c\in\mathcal{C}} \sum_{\bm{u}^{(c)}} \left[ \prod_{X\in\bm{X}_C} P(x|\mathrm{pa}_X)
\prod_{U\in \bm{U}_c} P(u)
\right]
=
\prod_{c\in\mathcal{C}} \prod_{X\in\bm{X}^{(c)}} \hat{P}(x|\bm{y}_X^{(c)})
\,.
\end{equation}
However, in real applications we shall always make use of finite samples. And the empirical distribution learnt from a finite sample may be arbitrarily far from the limit case, so as to make Eq.~\eqref{eq:compat} possibly fail. In this case, the empirical distribution is incompatible with the SCM under consideration, in the sense that the SCM simply cannot generate such a distribution in the limit. For this reason, we call Eq.~\eqref{eq:compat}   $M$-\emph{compatibility}, or compatibility of the data with a given model.

$M$-compatibility is particularly important when we are only given a PSCM. In this case, we aim at reconstructing the uncertainty about $\bm{U}$ from the empirical PMF $P(\bm{X})$. But if $M$-compatibility fails, we know that the task is hopeless, because there is no $P(\bm{U})$ that can eventually lead to $P(\bm{X})$. This may happen either because the sample is too small, or because the PSCM is a wrong model of the phenomenon under study. In either case, one should refrain from making inferences using jointly the PSCM and $P(\bm{X})$ as they would be logically contradicting each other.

We thus have to make sure that $M$-compatibility holds before computing inferences. To this end, let us consider the marginal log-likelihood of a given PSCM. By Eq.~\eqref{eq:joint2} we obtain:
\begin{equation}\label{eq:liku}
LL[\{P(U)\}_{U\in\bm{U}}] := 
\sum_{c\in\mathcal{C}} \sum_{\bm{y}^{(c)}} n(\bm{y}^{(c)}) \log \sum_{\bm{u}^{(c)}} \left[ \prod_{X\in\bm{X}^{(c)}} P(x|\mathrm{pa}_X) \prod_{U\in\bm{U}^{(c)}} P(u) \right]
\,,
\end{equation}
where by definition
$\bm{Y}^{(c)}=\cup_{X\in\bm{X}^{(c)}}\{ X,\bm{Y}^{(c)}_X\}$
and hence $n(\bm{y}^{(c)})=\sum_{X\in\bm{X}^{(u)}} n(x,\bm{y}_X^{(c)})$. The following result characterises the global maxima of the marginal likelihood:
\begin{thm}\label{th:unimod}
Let $\mathcal{K}$ denote the set of quantifications for $\{P(U)\}_{U\in\bm{U}}$ consistent with the following constraint to be satisfied for each $c\in\mathcal{C}$ and each $\bm{y}^{(c)}$:
\begin{equation}\label{eq:localconstraints}
\sum_{
\substack{\bm{u}^{(c)}: 
f_X(\mathrm{pa}_X)=x\\ \forall X\in\bm{X}^{(c)}}}
\prod_{U\in\bm{U}^c} P(u) =
\prod_{X\in\bm{X}^{(c)}}
\hat{P}(x|\bm{y}^{(c)}_X)
\,,
\end{equation}
where the values of $u$, $x$ and $\bm{y}^{(c)}_X$ are those consistent with $\bm{u}^{(c)}$ and $\bm{y}^{(c)}$. If $\mathcal{K} \neq \emptyset$, the log-likelihood in Eq.~\eqref{eq:liku} achieves its global maximum, equal to $LL^*$, if and only if  $\{P(U)\}_{U\in\bm{U}} \in \mathcal{K}$. If $\mathcal{K}=\emptyset$, the marginal log-likelihood in Eq.~\eqref{eq:liku} can only take values strictly lower than $LL^*$.
\end{thm}
In other words, global optimality of the marginal log-likelihood is tantamount to finding an FSCM compatible with the data and the given PSCM:
\begin{cor}\label{cor:k}
The function in Eq.~\eqref{eq:liku} achieving the global maximum of the marginal log-likelhood and having $\{P(U)\}_{U\in\bm{U}}\in\mathcal{K}$ are equivalent conditions for $M$-compatibility.
\end{cor}
This means that the marginal log-likelihood at its maximum height $LL^*$ looks like a flat plateau. Such a plateau characterises all and only the FSCMs compatible with the given PSCM. In case there are none, the maximum will be strictly lower than $LL^*$, which is something that we can use as an effective test of $M$-incompatibility.

\section{EM for Causal Computations}\label{sec:em}
With $\mathcal{K}$ available, one could compute the bounds of counterfactual queries by optimising the corresponding function of $\{P(U)\}_{U\in\bm{U}}$ over $\mathcal{K}$. An explicit characterisation of $\mathcal{K}$ has been provided for the special case of quasi-Markovian models in another paper \cite{zaffalon2020};  that approach faces computational issues if the model is non-Markovian, though. In fact coping with partial identifiability is demanding even with simple topologies and queries, as we show by the following:

\begin{thm}\label{th:hard}
Computing interventional queries in polytree-shaped PSCMs is NP-hard.
\end{thm}

This result makes it clear that we should generally consider approximate methods to compute bounds. Cor.~\ref{cor:k} provides a good match in this respect: it tells us that sampling the global optimum points of the marginal log-likelihood corresponds to sampling from, and hence approximating, $\mathcal{K}$.
And the crucial observation is that we can easily sample the optimum points of the marginal log-likelihood with the \emph{expectation-maximisation} (EM) scheme \cite{dempster1977maximum}: in fact, exogenous variables are missing at random in $\mathcal{D}$, just because they are latent (missing with probability one).

In particular, given an initialisation $\{P_0(U)\}_{U\in\bm{U}}$, the EM algorithm consists in regarding the posterior probability $P_0(u|\bm{x})$ as a \emph{pseudo-count} for $(u,\bm{x})$, for each $\bm{x}\in\mathcal{D}$,  $u\in\Omega_U$ and $U\in\bm{U}$ (E-step). A new estimate is consequently obtained as $P_1(u):=|\mathcal{D}|^{-1}\sum_{\bm{x}\in\mathcal{D}} P_0(u|\bm{x})$ (M-step). This scheme, called EMCC (\emph{EM for Causal Computation}) and detailed in Alg.~\ref{alg:cem}, is iterated until convergence. Subroutine ${\tt ccomponents}$ (line 2) finds the c-components of $M$, ${\tt initialisation}$ (line 1) provides a random initialisation of the exogenous PMFs, line 4 corresponds to a restriction of the data set to the variables in the $c$-th c-component (line 4), while $\varepsilon$ is the threshold to evaluate parameter convergence with respect to a metric $\delta$ (line 6). Note that in line 8 the d-separation properties of c-components allow to replace $P(U|\bm{x})$ with $P(U|\bm{y}^{(c)})$, where $c$ is such that $U\in\bm{U}^{(c)}$.

\begin{algorithm}[htp!]
\caption{EMCC: given SCM $M$ and data set $\mathcal{D}$ returns $\{P(U)\}_{U\in\bm{U}}$.}
\begin{algorithmic}[1]
\STATE $\{P_0(U)\}_{U\in\bm{U}} \leftarrow {\tt initialisation}(M)$
\STATE $\{\bm{U}^{(c)}, \bm{X}^{(c)}\}_{c\in\mathcal{C}} \leftarrow {\tt ccomponents}(M)$
\FOR{$c \in \mathcal{C}$}
\STATE $\mathcal{D}^{(c)} \leftarrow \mathcal{D}^{\downarrow \bm{Y}^{(c)}}$
\STATE $t \leftarrow 0$
\WHILE{$\delta[\{P_{t+1}(U)\}_{U\in\bm{U}},\{P_t(U)\}_{U\in\bm{U}}]> \varepsilon$}
\FOR{$U\in \bm{U}^{(c)}$}
\STATE $P_{t+1}(U) \leftarrow |\mathcal{D}|^{-1}\sum_{\bm{y}^{(c)} \in \mathcal{D}^{(c)}} P_t(U|\bm{y}^{(c)})$
\STATE $t \leftarrow t+1$

\ENDFOR
\ENDWHILE
\ENDFOR
\end{algorithmic}\label{alg:cem}
\end{algorithm}

And since $P_{t+1}$ gets a higher marginal log-likelihood than $P_t$ \cite[Thm.~19.3]{koller2009}, it readily follows that

\begin{cor}\label{cor:emcc}
EMCC always converges returning a point of $\mathcal{K}$.
\end{cor}
Thus multiple EMCC runs, started with different seeds, will yield an approximating subset of $\mathcal{K}$.

As a side remark, note that each iteration of the loop in line 3 of Alg.~\ref{alg:cem} can be executed in parallel to the others because of the d-separation among c-components. Something similar can be done at the data set level when computing, by standard BN algorithms, the queries in line 8. 

\section{\emph{M}-Compatibility and the Limits of Rung 2 of Pearl's Hierarchy}\label{sec:incomp}
In this section we give insights on $M$-compatibility, as defined in Sect.~\ref{sec:lik}, and on its consequences. Let us first consider a simple example to be used along the section to clarify our findings.

\begin{exe}[]\label{ex:pearl}
Consider the example in
\cite[Sect.~4.1]{exact2021causes}. It refers to a study about the recovery $Y$ of patients of gender $Z$ possibly subject to treatment $X$.
 
A sample of 700 patients is considered (the corresponding frequencies are in Tab.~\ref{tab:study}). The authors show that for an SCM whose graph is like in Fig.~\ref{fig:scmtoy}f, given the sample, the probability of necessity and sufficiency for treatment ($X$) on effect ($Y$) is no greater than $0.01$ (or PNS: $P(Y_{X=1}=1,Y_{X=0}=0)$ \cite{pearl2009causality}).
 They tell this for any SCM with that graph, irrespectively of the specific SEs it uses; they only require that the empirical distribution factorises according to the graph. 
\end{exe}

\vspace{-0.1cm}
\begin{table}[htp!]
	\centering
	{\scriptsize
		\begin{tabular}{cccr}
			\toprule
			Gender ($Z$) & Treatment ($X$)&Recovery ($Y$)&$\#$\\
			\midrule
			0&0&0&2\\
			0&0&1&114\\
			0&1&0&41\\
			0&1&1&313\\
			\bottomrule
	\end{tabular}\hspace{10pt}\begin{tabular}{cccr}
	\toprule
	Gender ($Z$) & Treatment ($X$)&Recovery ($Y$)&$\#$\\
	\midrule
	1&0&0&107\\
	1&0&1&13\\
	1&1&0&109\\
	1&1&1&1\\
	\bottomrule
\end{tabular}
}
\vspace{0.2cm}
\caption{Data from an observational study involving three Boolean variables \cite{exact2021causes}. Positive states means female ($Z$), treated ($X$) and recovered ($Y$).}
	\label{tab:study}
\end{table}
\vspace{-0.1cm}
Let us call \emph{conservative} an SCM whose SEs are specified in a conservative way. We focus on Markovian models because our notion of conservative SE, in Sect.~\ref{sec:background}, refers to this case only (an extension to general SMCs can be obtained relying on recent work \cite{duarte2021, zhang2021}). The following holds:

\begin{thm}\label{th:conservative}
Let $M$ be a conservative Markovian SCM over $(\bm{X},\bm{U})$. For any empirical $P(\bm{X})$, there exists at least a specification of $\{P(U)\}_{U\in\bm{U}}$ that guarantees $M$-compatibility.
\end{thm}

In practice, by Thm.~\ref{th:conservative}, if the SEs of a Markovian SCM are not available and we adopt a conservative specification, compatibility is guaranteed and, by Cor.~\ref{cor:emcc}, EMCC can  eventually be used for inference. This is what we do in the following example, which refers to the same setup of Ex.~\ref{ex:pearl}.

\begin{exe}\label{ex:pearl2}
Assume that the Markovian SCM in Fig.~\ref{fig:scmtoy}f has generated the data in Tab.~\ref{tab:study}. In particular assume $M$ conservative. This corresponds to $|\Omega_W|=2$, $|\Omega_U|=4$ and $|\Omega_V|=16$. 
For the SEs we have that $f_Z$ is just the identity map, while $f_X$ is such that:
$$f_X(Z,U=0)=\neg Z\,, f_X(Z,U=1)=0\,, f_X(Z,U=2)=Z\,, f_X(Z,U=3)=1\,.$$
The relations between $Y$ and $(X,Z)$ according to $f_Y$ and corresponding to the sixteen states of $V$ can be similarly listed. $M$-compatibility with data in Tab.~\ref{tab:study} corresponds to the constraints in Eq.~\eqref{eq:localconstraints}. For $W$, this trivially means $P(W=0)=P(Z=0)=\frac{470}{700}$ and $P(W=1)=P(Z=1)=\frac{230}{700}$. For $U$ we have instead the two following independent constraints for PMF $P(U)$:
\begin{eqnarray}\label{eq:c1}
P(U=1)+P(U=2) &=& P(X=0|Z=0)=\frac{116}{470}\,,\\ \label{eq:c2}
P(U=0)+P(U=1) &=& P(X=0|Z=1)=\frac{120}{230}\,.
\end{eqnarray}
Four independent linear constraints can similarly be obtained for $P(V)$. It is a straightforward exercise to obtain a parametrisation for all $P(U)$ and $P(V)$ specifications consistent with the above constraints, and verify that the corresponding interval for PNS is the same as the one given in \cite{exact2021causes} on the basis of bounding formulae. Note that in that paper the upper bound has been rounded to the second decimal place, while a more precise estimate would be $0.015$. This is the same number we get with EMCC if we compute PNS for the conservative specification of the SCM in Fig.~\ref{fig:scmtoy}f. 
\end{exe}

Thus the bounds in \cite{exact2021causes} correctly predict the right numbers we get in the above example. But we are not only assuming that the data factorise according to the graph in Fig.~\ref{fig:scmtoy}f, as the mentioned paper does. We are also assuming that the specification is conservative (i.e., all the mechanisms are possible). What happens if we do not---considered that the latter is not an assumption required to derive the bounds. This is discussed in the next example, still based on the setup of Ex.~\ref{ex:pearl}.

\begin{exe}\label{ex:pearl3}
Assume that, because of some expert knowledge, the state $U=3$ is impossible, this meaning that a deterministic mechanism forcing the treatment of all the patients is considered unrealistic. Such an assumption preserves the separate surjectivity of $f_X$ and also the joint surjectivity of the model. In other words, for each joint state of $(X,Y,Z)$, there is at least a joint state of $(W,U,V)$ inducing it. Under this additional information, the negation of Eq.~\eqref{eq:c1} becomes:
\begin{equation}\label{eq:c3}
P(U=0)=P(X=1|Z=0)=\frac{354}{470}
\,,
\end{equation}
and this is clearly inconsistent with Eq.~\eqref{eq:c2}. 
\end{exe}

This means that the model obtained by dropping state $U=3$ cannot, in the limit, generate the distribution that we see in the data: namely, the data are incompatible with such a non-conservative SCM. Note also that this shows that the empirical distribution of the data can factorise according to the graph, while at the same time, it may be incompatible with the related SCM. This may happen because the latter form of incompatibility depends on the SEs, not only on the graph.

In fact, as shown by the next example, $M$-incompatibility is subtle in that it may be based on an empirical probability that is inconsistent with the very SCM used to produce the data!

\begin{exe}\label{ex:pearl4}
In the same setup of Ex.~\ref{ex:pearl3} assume that some expert tells us that, out the sixteen states of $V$, only three states are possible, namely those indexing the following logical relations:
$$f_Y(X,Z,V=0)=X \vee \neg Z\,,
f_Y(X,Z,V=1)=\neg X \vee \neg Z\,,
f_Y(X,Z,V=2)=\neg X \wedge Z\,.$$
Note that this preserves surjectivity. Call $M'$ the reduced SCM that we obtain from the conservative specification (in $M$) by keeping only the values $0,1,2$ both for $U$ and $V$. Consider the following distributions for the exogenous variables in $M'$: $P(V)=[0.47, 0.439, 0.091]$, $P(U)=[0.677,0.000,0.323]$, and $P(W)$ taking the same empirical values of $P(Z)$ as in the case of $M$. Because of the previous discussion, the data in Tab.~\ref{tab:study} are clearly $M'$-incompatible. And yet they can be produced by $M'$, because $P(X,Y,Z)$ is a positive distribution under $M'$. So any data can be generated---with different log-likelihoods. In particular, the ratio between the marginal log-likelihood of $M'$ and that of $M$ is $0.71$: thus it turns out that it is not at all unlikely to produce the data in Tab.~\ref{tab:study} with model $M'$. And since those data factorise according to the graph, as before, we should be allowed to apply the bounds as before, claiming that the upper bound of PNS for $M'$ is $0.015$. But the PNS for $M'$ is $0.15$, 10 times above the bound. And so the bounds in \cite[Sect.~4.1]{exact2021causes} fail here.
\end{exe}

They fail because `factorising according to the graph' is not a strong enough assumption to derive bounds; we should ask for more, namely, $M$-compatibility with the underlying SCM. Stated differently, whenever we produce formulae, or algorithms, for computing counterfactuals, we should make sure that $M$-compatibility holds, otherwise the results will be unwarranted. Remember that $M$-compatibility can easily be verified in general through EMCC (see Cor.~1 and the paragraph that follows it); we are not aware or any other such simple and general method to verify it. 

At this point, one might be tempted to escape the problem of testing $M$-compatibility in absence of SEs (i.e., in presence of data and the causal graph alone) by using a conservative specification of the SEs, given that the resulting model is always compatible with the data (cf. Thm.~\ref{th:conservative}). Even more so that recent efforts appear to provide a conservative specification for general SCMs \cite{duarte2021, zhang2021}; this would seem to enable doing general counterfactual inference without SEs. Yet, conservative specifications do not seem to provide us with an easy way out, as shown next.

Let us recall that the notion of $M$-compatibility in Eq.~\eqref{eq:compat} refers to FSCMs. For PSCMs, we should consider the set $\mathcal{K}$ as in Thm.~\ref{th:unimod}. Each element of $\mathcal{K}$ defines an FSCM, which is $M$-compatible by construction. Whence $M$-compatibility with a PSCM  corresponds to  $\mathcal{K}$ being non-empty.

Note that with Markovian (and quasi-Markovian) models, the constraints in Eq.~\eqref{eq:localconstraints} can be written separately for each $U\in\bm{U}$; we denote by $\mathcal{K}(U)$ the set of specifications of $P(U)$ consistent with those `local' constraints. In this respect, $\mathcal{K}$ can be understood as the set of stochastic products of elements from sets $\mathcal{K}(U)$, with $U\in\bm{U}$. 

Given a Markovian conservative PSCM $M$, any other Markovian PSCM $M'$ over the same endogenous variables and graph can be defined by dropping some values of the exogenous variables, along with their corresponding SEs. That is, $M'$ is defined via sets $\{\Omega_U'\}_{U\in\bm{U}}$ such that, for each $U\in\bm{U}$, $\Omega_U'\subseteq\Omega_U$ are the states of $U$ indexing the deterministic relations of the corresponding SE of $M'$. As a consequence we can regard a conservative PSCM as the set of all the possible non-conservative PSCMs. Some of these may be incompatible with the data, though. So how does inference in the conservative PSCM and in the set of non-conservative PSCMs relate to each other?   

To answer this question, let us say that $M$ \emph{embeds} $M'$ if and only if $M$ gives a chance to $M'$ to be the actual model underlying the data, i.e., if for all $U\in\bm{U}$ there is at least a $P(U)\in \mathcal{K}(U)$ so that $P(U \in \Omega_U')>0$. The following holds:
\begin{thm}\label{cor:discarding}
A Markovian conservative PSCM $M$ cannot embed incompatible models.
\end{thm}

As a consequence, running EMCC, or, e.g., the inference algorithms in \cite{zaffalon2020} in a conservative model corresponds to \emph{automatically discarding} the incompatible models while using only the compatible ones. The important implication here is that if the true underlying model is not compatible with the available data, the results obtained by using the conservative model will be unwarranted as an approximation to the actual one. This is in fact the reason why in Example~\ref{ex:pearl4} the PNS interval obtained by the conservative model does not contain the actual value of PNS.

The lesson appears to be that we cannot have guaranteed bounds without knowing the SEs of the underlying SCM. Using the conservative specification as a replacement for the actual SEs might help under the assumption that the sample is large enough so as it becomes $M$-compatible with the true, underlying, model. The challenge is to render such an assumption tenable. 

\section{Evaluating the EMCC}\label{sec:exp}
We start by characterising the accuracy of our procedure in terms of credible intervals. 
\begin{thm}\label{th:credible}
Let $[a^*,b^*]$ denote the exact probability bounds of a causal query. Say that $\rho:=\{r_i\}_{i=1}^n$ are the outputs of $n$ EMCC iterations, while $[a,b]$ is the interval induced by $\rho$, i.e., $a:=\min_{i=1}^n r_i$ and
$b:=\max_{i=1}^n r_i$.
By construction $a^*\leq a \leq b \leq b^*$. 
The following inequality  holds:
\begin{equation}\label{eq:ab}
P\left(
a-\varepsilon L\leq
a^*\leq b^* \leq b+\varepsilon L\,\bigg|\, \rho \right) =
\frac{
1+(1+2\varepsilon)^{2-n}-2(1+\varepsilon)^{2-n}}{
(1-L^{n-2})
-(n-2)(1-L)L^{n-2}}
\,,
\end{equation}
where $L:=(b-a)$ and $\varepsilon:=\delta/(2L)$ is the relative error at each extreme of the interval obtained as a function of the absolute allowed error $\delta\in(0,L)$.
\end{thm}
Eq.~\eqref{eq:ab} can be used to evaluate the EMCC accuracy in approximating the true interval $[a^*,b^*]$. For example, if we allow for a maximum 17\% of error ($\varepsilon$) at each extreme of the interval $[a,b]$, 20 EM runs achieve 95\% credibility. By numerically taking the limit of Eq.~\eqref{eq:ab} with $L$ that tends to the machine zero, we can address the case of (candidate) identifiable queries, i.e., $n$ EMCC runs always returning the same value. In this case 6 runs achieve 95\% credibility; 9 runs lead to 99\%.

We also evaluate the EMCC via empirical experiments. This will provide us with an actual measure of its performance. A benchmark of 460 SCMs is considered. These are Boolean endogenous chains of $m\in\{$5,7,10$\}$ nodes augmented with: an exogenous parent per node in the Markovian case, exogenous parents of pairs of randomly picked endogenous nodes whose distance in the chain is no more than two in the other cases. A single exogenous node per c-component is allowed in the quasi-Markovian case, two in non-quasi-Markovian models.\footnote{The EMCC code is part of the CREDICI library \cite{cabanas2020a}. The code to reproduce the models and the experiments is available at \url{https://github.com/IDSIA-papers/2021-NeurIPSWHY-causalEM}.
}

\begin{figure}[htp!]
\centering
	\begin{tikzpicture}[scale=.6]
	\begin{groupplot}[group style={group size=3 by 1, vertical sep=0cm, horizontal sep=1.7cm}, height=5cm,width=7.5cm]
	\nextgroupplot[
	tick align=outside,
	tick pos=left,
	title={$\;\;\;\;\;\;\;\;\;\;$Markovian ($|\bm{X}^{(c)}|=|\bm{U}^{(c)}|=1$)},
	x grid style={white!69.0196078431373!black},
	xlabel={EMCC Runs ($n$)},
	xmajorgrids,
	xmin=0.0, xmax=21.05,
	xtick style={color=black},
	y grid style={white!69.0196078431373!black},
	ylabel={RMSE},
	ymajorgrids,
	ymin=-0.002, ymax=0.032,
	ytick={0.00, 0.01, 0.02, 0.03},
    yticklabels={$0.00$, $0.01$, $0.02$, $0.03$},
    ytick scale label code/.code={},
	ytick style={color=black}]
	\addplot [black,ultra thick]
	table {%
1	0.012376214838653938
2	0.008422817468105252
3	0.008063445788194136
4	0.007876810676310266
5	0.007618002043117809
6	0.007362722005726768
7	0.007307494069198166
8	0.007364657328206295
9	0.007123672565124886
10	0.007029855696506702
11	0.005817412033287176
12	0.005797403256420669
13	0.005877158632384475
14	0.0057166366691337655
15	0.005716636657822749
16	0.005687690622908202
17	0.005687690622908202
18	0.005666381635865105
19	0.005545431402151152
20	0.005462140691054489
	};
	\addplot [black,dashed,ultra thick]
	table {
1	0.002007783155483807
2	0.0015350221611361973
3	0.001502352867542255
4	0.0012664408990718946
5	0.001265535510154942
6	0.0011508764301720116
7	0.0011508748076437933
8	0.0011508534555897125
9	0.0010970634203283293
10	0.0008866431076642294
11	0.0008876053496098802
12	0.000888174085460758
13	0.000885674418816315
14	0.0008876265047378738
15	0.0008787508268354913
16	0.0008787519180302644
17	0.0008787519180302644
18	0.0008828996719234623
19	0.0008834529747060114
20	0.0008836094232607001
	};
	\addplot [black,dotted,ultra thick]
	table {%
1	0.0011963828419024147
2	0.0010183391955714365
3	0.000883414737248363
4	0.000879939117831595
5	0.000871478604452136
6	0.0008213845266914128
7	0.0006921351714967062
8	0.0004030623709317652
9	0.00040142088681647343
10	0.00040144041772452465
11	0.0003857909622926413
12	0.0003757078618728462
13	0.0002794278455499331
14	0.0002794924800680011
15	0.0002795780249468233
16	0.0002795780168268272
17	0.0002684889042419701
18	0.00026687385243242986
19	0.0002668160010864103
20	0.0002668196204089061
	};
	
	\nextgroupplot[
	scaled y ticks=manual:{}{\pgfmathparse{#1}},
	tick align=outside,
	tick pos=left,
	title={Quasi-Markovian ($|\bm{U}^{(c)}|=1$)},
	x grid style={white!69.0196078431373!black},
xlabel={EMCC Runs ($n$)},
xmajorgrids,
xmin=0.0, xmax=21.05,
xtick style={color=black},
y grid style={white!69.0196078431373!black},
ymajorgrids,
ymin=-0.002, ymax=0.032,
ytick={0.00, 0.01, 0.02, 0.03},
yticklabels={$0.00$, $0.01$, $0.02$, $0.03$},
ytick style={color=black}
	]
	\addplot [black,ultra thick]
	table {
1	0.02434804011643076
2	0.015480640547975134
3	0.005617951948300049
4	0.005639507330188557
5	0.005202631757763924
6	0.004764296357539729
7	0.004574724589211165
8	0.004534190743996103
9	0.004522948985655133
10	0.004514971382570055
11	0.004466478126712918
12	0.004508154003190278
13	0.004508154003190278
14	0.0044758371391453436
15	0.004479394172869698
16	0.004460717126272496
17	0.004417901691879873
18	0.004417901691879873
19	0.004344853621463296
20	0.004249459101929033
	};
		\addplot [black,dashed,ultra thick]
	table {
1	0.018034032624730672
2	0.01667418501796891
3	0.01568745558171788
4	0.008705591416606663
5	0.008705591416606663
6	0.00783669117363892
7	0.007828901704006299
8	0.007831062787358838
9	0.007829307208214117
10	0.007807185269950675
11	0.007802800358739968
12	0.007802800358739968
13	0.007802800358739968
14	0.007802800358739968
15	0.007798634624293953
16	0.007798634624293953
17	0.007798634624293953
18	0.007798634624293953
19	0.007798508221595275
20	0.007798508221595275
	};
	\addplot [black,dotted,ultra thick]
	table {
1	1.442399202406283e-05
2	1.4197366902837842e-05
3	1.483855211456964e-05
4	1.483855211456964e-05
5	1.483870280390128e-05
6	1.483870280390128e-05
7	1.461387784665595e-05
8	1.4613878064328965e-05
9	1.4617099012917957e-05
10	1.468642929953879e-05
11	1.468676429754995e-05
12	1.468676429754995e-05
13	1.4515173823457624e-05
14	1.4515173823457624e-05
15	1.4515173862954448e-05
16	1.4515173862954448e-05
17	1.4104102410099289e-05
18	1.4104102410099289e-05
19	1.4106245273413472e-05
20	1.4106245273413472e-05
	};
	
	\nextgroupplot[
	legend cell align={left},
	legend pos=north east,
	legend style={fill opacity=0.8, draw opacity=1, text opacity=1, draw=white!80!black},
	scaled y ticks=manual:{}{\pgfmathparse{#1}},
	tick align=outside,
	tick pos=left,
	title={Non-quasi-Markovian ($|\bm{U}^{(c)}|=2$)},
	x grid style={white!69.0196078431373!black},
xlabel={EMCC Runs ($n$)},
xmajorgrids,
xmin=0.0, xmax=21.05,
xtick style={color=black},
y grid style={white!69.0196078431373!black},
ymajorgrids,
	ymin=-0.002, ymax=0.032,
	ytick={0.00, 0.01, 0.02, 0.03},
yticklabels={$0.00$, $0.01$, $0.02$, $0.03$},
ytick style={color=black}
	]
	\addplot [ultra thick,black]
	table {
1	0.030827921062741827
2	0.02081755145957289
3	0.01667378697682377
4	0.012815063419875544
5	0.010771120121313253
6	0.010235032368453493
7	0.008067719787796936
8	0.007947782689461145
9	0.007679940702231823
10	0.007163263444547301
11	0.0068359251382264135
12	0.006684888012115456
13	0.006359886734970164
14	0.005727400752336488
15	0.005252887055498689
16	0.005252887052567559
17	0.005130779699332728
18	0.005081721225325334
19	0.005042970768671365
20	0.005042970768671365
21	0.004375532175605437
22	0.004374229510879603
23	0.002403809102340001
24	0.0024037460459983354
25	0.001921726809697233
26	0.0018365873002905654
27	0.0018043699925708854
28	0.0017150710111219051
29	0.001667246190855419
30	0.0
	};\addlegendentry{$m=5$}

	\addplot [black,dashed,ultra thick]
	table {
1	0.00462563625553438
2	0.002835100300716107
3	0.0022961344503695962
4	0.0021188931145448803
5	0.0018565208537950054
6	0.0014977483897356382
7	0.0014596908536131982
8	0.0013208108430096822
9	0.0012472171493687306
10	0.0008539428915857111
11	0.0008062364551592794
12	0.0005765259050890833
13	0.00036629323579520445
14	0.00031713414927435536
15	0.00030052404557139305
16	0.00030052404557139305
17	0.00026311330085264405
18	0.00025381387110202686
19	0.00025259881623154187
20	0.00024377620874628595
21	0.000238368162321741
22	0.00020053349131580024
23	0.000200530410530493
24	0.00016953737318702332
25	0.00016946133149879
26	0.00016946133149879
27	0.00016946133149879
28	0.00015434876292391104
29	0.00015365022450647024
30	0.0
	};
	\addlegendentry{$m=7$}
	\addplot [black,dotted,ultra thick]
	table {%
1	0.0006434542443909193
2	0.0002570559776203546
3	0.00021382356212325615
4	0.00020883083728865423
5	0.0001980303452573439
6	0.00018701810150835226
7	0.00018169980633809274
8	0.00018055478799557518
9	0.00018050919473611298
10	0.0001508649636692269
11	0.00014302292410930947
12	0.00010844557516176569
13	8.719232611815989e-05
14	8.716035396828703e-05
15	8.102546487997393e-05
16	8.03706964128335e-05
17	5.358951225618521e-05
18	5.2483381667374624e-05
19	5.171236479827636e-05
20	5.0557447725399626e-05
21	4.8438300449921127e-05
22	4.8105751473604875e-05
23	4.6448684487328396e-05
24	4.6448684487328396e-05
25	4.541641693572381e-05
26	4.300435694213211e-05
27	4.300435694213211e-05
28	4.2146994663111654e-05
29	0.0
30	0.0
	};
	\addlegendentry{$m=10$}
	\end{groupplot}
	\end{tikzpicture}
	\vspace{-5pt}
	\caption{EMCC RMSE vs. number of runs ($n$).}	\label{fig:rmse}
\end{figure}
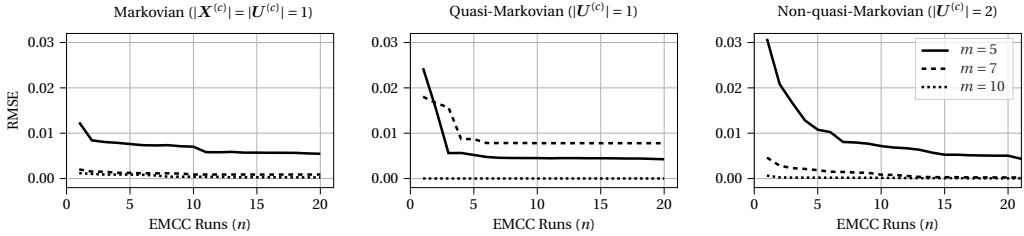

We consider PNS queries involving the first and the last endogenous variable. We start from a `true' FSCM $M$, from which a data set $\mathcal{D}$ of $1000$ instances is sampled. Afterwards the corresponding PSCM is considered. Endogenous data are processed by EMCC with random initialisation, KL divergence with double machine epsilon decides convergence (Alg.~\ref{alg:cem}, line 6). PNS queries are addressed in the derived counterfactual graph and the bounds with respect to multiple EMCC runs evaluated. The baseline procedure in \cite{zaffalon2020} allows one to compute the exact bounds for Markovian and quasi-Markovian models. For the non-quasi-Markovian case, exact bounds are unavailable; for the sake of evaluation, we assume they coincide with those we get by a very large number of EMCC runs. Fig.~\ref{fig:rmse} depicts, as a function of the number of the EM runs, the average RMSE for the approximate bounds with respect to the baseline (i.e., $\sqrt{((a^*-a)^2+(b^*-b)^2)/2}$).

The behaviour is consistent with Thm.~\ref{th:credible}: the EMCC gives an inner approximation that quickly ($n\le20$) converges to a very accurate estimate ($\mathrm{RMSE}<0.01$). For small PNS intervals (e.g., for $m=10$ in non-quasi-Markovian models), ten runs allow to span the exact interval, again quite consistently with Thm.~\ref{th:credible}. The case $m=10$ for quasi-Markovian models is even more extreme as the exact PNS interval is small and numerically collapses into a single sharp value. In such cases, by construction, the EMCC gives the exact value in a single run. 

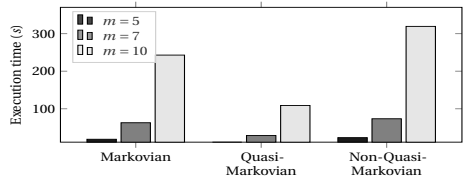
\begin{figure}[htp!]
\centering
\begin{tikzpicture}[yscale=.5,xscale=0.6]
\begin{axis}[
ybar=3pt,
enlarge x limits=0.3,
enlarge y limits=0.00,
ymax=370,
legend cell align={left},
legend pos=north west,
legend style={fill opacity=0.8, draw opacity=1, text opacity=1, draw=white!80!black},
ylabel={Execution time ($s$)},
symbolic x coords={Markovian,Quasi-Markovian,Non-Quasi-Markovian},
xtick=data,
x tick label style={rotate=0, text width=2cm,align=center},
nodes near coords align={vertical},
every node near coord/.append style={font=\tiny},
bar width=0.65cm,
width=0.75\textwidth,
height=.37\textwidth,
]
\addplot[draw = black,fill = black!90]  coordinates {
(Markovian,18.8) (Non-Quasi-Markovian,23.1) (Quasi-Markovian,11.1)};
\addplot[draw = black,fill = black!50]  coordinates {
(Markovian,62.8) (Non-Quasi-Markovian,73.4) (Quasi-Markovian,28.69)};
\addplot[draw = black,fill = black!10]  coordinates {
(Markovian,242.9) (Non-Quasi-Markovian,319.3) (Quasi-Markovian,108.8)};
\legend{$m=5$,$m=7$,$m=10$}
\end{axis}
\end{tikzpicture}
\caption{Average EMCC execution times.}
\label{fig:interval_size}
\end{figure}
\begin{figure}[htp!]
\centering
\hspace{-3pt}
\begin{tikzpicture}[yscale=.5,xscale=0.6]
\begin{axis}[
ybar=3pt,
enlarge x limits=0.3,
enlarge y limits=0.0,
ymax=0.065,
legend cell align={left},
legend pos=north east,
legend style={fill opacity=0.8, draw opacity=1, text opacity=1, draw=white!80!black},
ylabel={Interval size},
symbolic x coords={Markovian,Quasi-Markovian,Non-Quasi-Markovian},
xtick=data,
ytick={0.00, 0.02, 0.04, 0.06},
yticklabels={$0.00$, $0.02$, $0.04$, $0.06$},
ytick scale label code/.code={},
x tick label style={rotate=0, text width=2cm,align=center},
every node near coord/.append style={font=\tiny},
bar width=0.65cm,
width=0.75\textwidth,
height=.37\textwidth,
]
\addplot[draw = black,fill = black!90] coordinates {
(Markovian,0.016) (Non-Quasi-Markovian,0.054) (Quasi-Markovian,0.033)};
\addplot[draw = black,fill = black!50] coordinates {
(Markovian,0.032) (Non-Quasi-Markovian,0.008) (Quasi-Markovian,0.034)};
\addplot[draw = black,fill = black!10] coordinates {
(Markovian,0.002) (Non-Quasi-Markovian,0.001) (Quasi-Markovian,0.00002)};
\legend{$m=5$,$m=7$,$m=10$}
\end{axis}
\end{tikzpicture}
\caption{Average interval sizes for PNS queries.}
\label{fig:interval_time}
\end{figure}
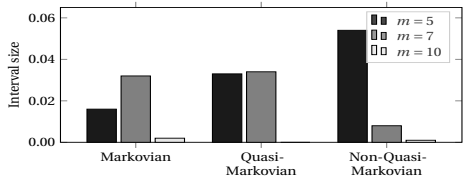

Fig.~\ref{fig:interval_size} depicts the average sizes of the PNS exact intervals. Fig.~\ref{fig:interval_time} shows the EMCC running times normalised with respect to the number of runs. The most time-consuming tasks are those involving non-quasi-Markovian models. This is expected as the complexity of our procedure is bounded by the treewidth of the reduced graph (Alg. \ref{alg:cem}, line 8), which is one for both Markovian and quasi-Markovian models. Note that quasi-Markovian models are presumably faster because of the lower number of exogenous nodes. In fact, $m$ only reflects the number of endogenous variables, the overall number of nodes is $2m$ in the Markovian case and $1.5m$ in the other cases. The PNS query is finally addressed in the counterfactual graph, whose dimension is $3m$ in the Markovian case, and $2.5m$ in other cases.

\section{Conclusions}\label{sec:conc}
We have presented a causal version of the EM algorithm that can be used to do counterfactual inference of unidentifiable queries. The algorithm is simple: its two components are the EM and some algorithms for BNs. It is an approximate algorithm too. This is a limitation. Yet, we have shown here that causal inference is NP-hard even in simple models, whence approximations seem necessary. Moreover, the EMCC is shown to offer a good compromise between efficiency and accuracy. In addition, EMCC gives us the opportunity to test the compatibility between a structural causal model and a sample. Checking it appears necessary to avoid drawing unwarranted conclusions. Future work should consider extending EMCC to continuous domains.
\appendix

\section{Proofs}\label{sec:app}
\begin{prf}
Let us first consider the case $\mathcal{K} \neq \emptyset$ and take $\{ P'(U)\}_{U\in \bm{U}}\in\mathcal{K}$. By definition, Eq.~\eqref{eq:localconstraints} for SCM $M$ rewrites as:
\begin{equation}\label{eq:localconstraints2}
\sum_{\bm{u}^{(c)}}
\left[
\prod_{X\in\bm{X}^{(c)}} P(x|\mathrm{pa}_X) \prod_{U\in\bm{U}^{(c)}} P'(u)
\right]=
\prod_{X\in\bm{X}^{(c)}}
\hat{P}(x|\bm{y}^{(c)}_X)
\,.
\end{equation}
Putting this in Eq.~\eqref{eq:liku} we get:
\begin{equation}\label{eq:likdec2}
LL = \sum_{c\in\mathcal{C}} \sum_{\bm{y}^{(c)}} n(\bm{y}^{(c)}) 
\sum_{X\in\bm{X}^{(c)}}
\log \hat{P}(x|\bm{y}_X^{(c)})\,,
\end{equation}
and this is Eq.~\eqref{eq:likdec} when the probabilities are those in Eq.~\eqref{eq:freqs}. This proves the sufficient condition. To prove the necessary condition, consider again Eq.~\eqref{eq:localconstraints2}, which is equivalent to Eq.~\eqref{eq:localconstraints}. Assume, \emph{ad absurdum}, that there is a $\{P'(U)\}_{U\in\bm{U}}$ where the log-likelihood attains its global maximum, but $\{P'(U)\}_{U\in\bm{U}} \not\in \mathcal{K}$. Thus, Eq.~\eqref{eq:localconstraints2} should be violated for at least a value of $\bm{y}^{(c)}$ and $c$. Yet, putting Eq.~\eqref{eq:localconstraints2} in Eq.~\eqref{eq:likdec} produces a value smaller than the global maximum, as this is achieved if and only if the values in Eq.~\eqref{eq:freqs} are used. The same discussion applies to any 
$\{P'(U)\}_{U\in\bm{U}}$ when $\mathcal{K}=\emptyset$. \qed
\end{prf}

\begin{prf}
The proof follows from the analogous result for credal networks in \cite[Thm.~1]{cozman2002}. The polytree used in that proof has deterministic conditional probability tables for non-root nodes. We can therefore understand that model as an SCM. Its endogenous nodes form a chain, each node has a single exogenous parent apart from the first one in the chain that has a second exogenous parent. The set-valued quantification of the exogenous PMFs corresponds to a partially specified SCM. As a query, the authors consider the upper bound of a marginal query in the last node of the chain. The task amounts to the identification of the upper bound of the causal effect on the last node of the chain given an intervention in an additional endogenous parent of the first node of the chain. An algorithm to bound interventional queries in SCMs would therefore solve inference in polytree-shaped credal networks. This contradicts the result of the authors. \qed
\end{prf}

\begin{prf}
The thesis is equivalent to the fact that the constraints in Eq.~\eqref{eq:localconstraints} can be satisfied for at least a $P(\bm{U})$. As $M$ is Markovian, for each $c\in\mathcal{C}$,
$|\bm{U}_c|=|\bm{X}_c|=1$ and we denote as $X$ and $U$ the unique elements of $\bm{X}_c$ and $\bm{U}_c$, while $\bm{Y}$ is the joint variable corresponding to the endogenous parents of $X$. Eq.~\eqref{eq:localconstraints} rewrites as the following linear constraint:
\begin{equation}\label{eq:linsys}
\sum_{u \in \Omega_U} P(x|\bm{y},u) \cdot P(u) = 
\sum_{u \in \Omega_U:f(\bm{y},u)=x} P(u) =
P(x|\bm{y})\,,
\end{equation}
to be satisfied for each $x\in\Omega_X$ and $\bm{y}\in\Omega_{\bm{Y}}$. The representability result provided in \cite[Thm.~1]{druzdzel1993causality} can be used to prove that the linear constraints in Eq.~\eqref{eq:linsys} can be satisfied. In their proof the authors consider a setup analogous to the current one but, instead of $U$, a continuous $U'\in[0,1]$ with a uniform density $P(U')$ is considered. Say that $\Omega_X := \{x_1,\ldots,x_q\}$. For each $\bm{y}\in\Omega_{\bm{Y}}$, we define the vector $H_{\bm{y}}:=\{h_{\bm{y}}^{(i)}\}_{i=0}^{q}$ such that $h_{\bm{y}}^{(0)}:=0$ and $h_{\bm{y}}^{(i)}:= \sum_{j=1}^i P(x_j|\bm{y})$ for each $i=1,\ldots,q$.
Note that $h_{\bm{y}}^{(i)}\leq h_{\bm{y}}^{(i+1)}$ for each $i=1,\ldots,q-1$ and $h_{\bm{y}}^{(q)}=1$. The authors show that Eq.~\eqref{eq:linsys} is satisfied if the SE $X=f_X'(\bm{Y},U')$ is such that:
\begin{equation}\label{eq:fprime}
f_X'(\bm{y},u') := \left\{ x_i \in \Omega_X : u' \in \left[
h_{\bm{y}}^{(i)},h_{\bm{y}}^{(i+1)} \right] \right\}\,,
\end{equation}
for each $u' \in [0,1]$ and $\bm{y}\in\Omega_{\bm{Y}}$. 
A partition of $[0,1]$ is obtained by
removing the left endpoints from the intervals in Eq.~\eqref{eq:fprime}  apart from the first one. Thus, for a given $\bm{y}\in\Omega_{\bm{Y}}$, Eq.~\eqref{eq:fprime} can be regarded as a discrete SE, mapping to $X$ the values of a discretisation of $U'$ based on the partition of $[0,1]$ induced by $H_{\bm{y}}$. 

A \emph{least common partition} is obtained from a set of partitions by simply putting together and sorting the endpoints of the intervals of all the partitions. Take the discretisation of $U'$ induced by such a least common partition when considering 
all the partitions of $[0,1]$ induced by each $\bm{y} \in \Omega_{\bm{Y}}$.
In practice, the SE $X=f_X'(\bm{Y},U')$ can be equivalently described by $X=\hat{f}_X(\bm{Y},\hat{U})$, where $\hat{U}$ is a discrete variable whose states are in correspondence with the above considered set of discretisation intervals for $U$. The uniform density $P(U')$ is consequently mapped to a PMF $P(\hat{U})$ such that $P(\hat{U}=\hat{u})$ is equal to the integral of $P(U')$ on the interval associated with $\hat{u}$ and hence it is equal to its width. Finally, observe that, for each $\hat{u}\in\Omega_{\hat{U}}$, $\hat{f}_X$ defines a deterministic relation between $\bm{Y}$ and $\bm{X}$ and this should correspond to a state of $U$ in $M$, as $U$ is enumerating all these possible relations because of the conservative specification. This defines a map $\mu:\Omega_{\hat{U}}\rightarrow\Omega_U$ with $\mu(\hat{u}):=\{ u \in \Omega_U: f_X(u,\bm{Y}) = \hat{f}(\hat{u},\bm{Y}) \}$. For $P(U)$ we have:
\begin{equation}
P(u) = \sum_{\hat{u} \in \Omega_{\hat{U}}: \gamma(\hat{u})=u} P(\hat{u})\,,
\end{equation}
and $P(u)=0$ for the states of $u$ that are not in the domain of $\mu$. This is the exogenous quantification that proves the thesis. \qed
\end{prf}

\begin{prf}
Assume, \emph{ad absurdum}, that $M$ embeds an incompatible $M'$. For each $U\in\bm{U}$, let $\{\mathcal{K}'(U)\}_{U\in\bm{U}}$ denote the sets of compatible specifications of $M'$. By Cor.~\ref{cor:k}, the incompatibility of $M'$ implies that $\mathcal{K}'(U)$ should be empty for at least a $U\in\bm{U}$. By the definition of embedding, for each $U\in\bm{U}$, we should have at least a $P(U)\in \mathcal{K}(U)$ such that $P(U \in \Omega_U')>0$.

Obtaining $M'$ from $M$ can be regarded as the result of conditioning the exogenous PMFs on the events $U\in\Omega_U'$, for each $U\in\bm{U}$. As $P(U \in \Omega_U')>0$, such conditioning is well defined for $P$. Let $P'(U)$ denote the resulting PMF for $M'$. 

Consider the compatibility constraints of $M$ as in Eq.~\eqref{eq:localconstraints} involving PMF $P(U)$. 
As $P(U) \in \mathcal{K}(U)$, these constraints should be satisfied by $P(U)$. With Markovian models, the constraints are linear. Dividing by $P(U \in \Omega_U')$ on both sides gives the analogous constraints for $P'(U)$. But this means $P'(U) \in \mathcal{K}'(U)$ and hence $\mathcal{K}'(U) \neq \emptyset$ (for each $U\in\bm{U}$), which is a contradiction.\qed
\end{prf}

\begin{prf}
Consider the probability on the left-hand side of Eq.~\eqref{eq:ab}.
The corresponding joint probability rewrites as:
\begin{equation}\label{eq:j0}
P\left(
\Delta_a \leq \frac{\delta}{2}
,\Delta_b \leq \frac{\delta}{2}
,\rho\right)=
\gamma \int\limits_{0}^{\frac{\delta}{2}}
\int\limits_{0}^{\frac{\delta}{2}}
P(\rho|\Delta_a=x,\Delta_b=y)
\mathrm{d}x
\mathrm{d}y\,,
\end{equation}
where a uniform prior density is considered, i.e., $P(\Delta_a=x,\Delta_b=y):=\gamma \mathrm{d}x \mathrm{d}y$ with $\gamma$ normalisation constant, $\Delta_a:=(a-a^*)$ and $\Delta_b:=(b^*-b)$. 

To obtain $P(\rho|\Delta_a = x,\Delta_b = y)$, let us first notice that, for each $r_i \in \rho$:
\begin{equation}\label{eq:rx}
P(r_i \in [a,b]|\Delta_a = x,\Delta_b = y) = \frac{b-a}{b-a+x+y}\,.
\end{equation}
Eq.~\eqref{eq:rx} is obtained by assuming the output $r_i$ of a single EMCC run to be uniformly distributed over the interval $[a^*,b^*]$. Under i.i.d. assumptions we get
$P(\rho|\Delta_a =x,\Delta_b=y)=(\frac{b-a}{b-a+x+y})^n$ from Eq.~\eqref{eq:rx}. Eq.~\eqref{eq:j0} rewrites therefore as:
\begin{equation}\label{eq:j}
P
\left(
\rho,\Delta_a \leq \frac{\delta}{2},\Delta_b \leq \frac{\delta}{2}
\right) = \gamma \int\limits_{0}^{\frac{\delta}{2}} 
\int\limits_{0}^{\frac{\delta}{2}} 
\left(\frac{b-a}{b-a+x+y}
\right)^{n}
\mathrm{d}x 
\mathrm{d}y 
\,.
\end{equation}
We can similarly obtain the marginal for $\rho$ by integrating $\Delta_a$ and $\Delta_b$, i.e.,
\begin{equation}\label{eq:m}
P(\rho) = \gamma \int\limits_{0}^{a+(1-b)} 
\left[
\int\limits_{0}^{a+(1-b)-y} 
\left(\frac{b-a}{b-a+x+y}
\right)^{n}
\mathrm{d}x
\right]
\mathrm{d}y
\,.
\end{equation}
The integrals in Eqs. \eqref{eq:j} and \eqref{eq:m} are trivial and their ratio gives the right-hand side of Eq.~\eqref{eq:ab}.\qed 
\end{prf}
\pagebreak
\bibliographystyle{plain}
\bibliography{emcc}
\end{document}